\documentclass[]{spie}  

\usepackage{amsmath,amsfonts,amssymb}
\usepackage{subcaption}
\usepackage{graphicx}
\usepackage{cite} 
\usepackage{epsfig}
\usepackage{nccmath}
\usepackage{xcolor,colortbl}
\usepackage{tabularx}
\usepackage{relsize}
\usepackage{pifont}
\usepackage{booktabs} 
\usepackage{multirow}
\usepackage{multicol}
\usepackage{adjustbox}
\usepackage{float}
\usepackage{makecell}
\usepackage{tabu}
\usepackage[colorlinks=true, allcolors=blue]{hyperref}
\usepackage[capitalize]{cleveref}
\usepackage{algorithm}
\usepackage{ulem}

\title{An Interpretable Vision Transformer as a Fingerprint-Based Diagnostic Aid for Kabuki and Wiedemann-Steiner Syndromes}

\author[a]{Marilyn Lionts}
\author[b]{Arnhildur Tomasdottir}
\author[c]{Viktor I. Agustsson}
\author[a]{Yuankai Huo}
\author[b, c, d]{Hans T. Bjornsson}
\author[e]{Lotta M. Ellingsen}

\affil[a]{Dept. of Computer Science, Vanderbilt University, Nashville, TN, USA}
\affil[b]{Dept. of Genetics and Molecular Medicine, Landspitali University Hospital, Reykjavik, Iceland}
\affil[c]{Louma G. Laboratory of Epigenetics Research, Faculty of Medicine, University of Iceland, Reykjavik, Iceland}
\affil[d]{McKusick-Nathans Dept. of Genetic Medicine, Johns Hopkins University School of Medicine, Baltimore, MD, USA}
\affil[e]{Faculty of Electrical and Computer Engineering  University of Iceland, Reykjavik, Iceland}

\authorinfo{Further author information: (Send correspondence to Marilyn Lionts)\\
E-mail: marilyn.m.lionts@vanderbilt.edu}

\begin{document}
\maketitle

\begin{abstract}
Kabuki syndrome (KS) and Wiedemann-Steiner syndrome (WSS) are rare but distinct developmental disorders that share overlapping clinical features, including neurodevelopmental delay, growth restriction, and persistent fetal fingertip pads. While genetic testing remains the diagnostic gold standard, many individuals with KS or WSS remain undiagnosed due to barriers in access to both genetic testing and expertise. Dermatoglyphic anomalies, despite being established hallmarks of several genetic syndromes, remain an underutilized diagnostic signal in the era of molecular testing.
This study presents a vision transformer-based deep learning model that leverages fingerprint images to distinguish individuals with KS and WSS from unaffected controls and from one another. We evaluate model performance across three binary classification tasks. Across the three classification tasks, the model achieved AUC scores of 0.80 (control vs. KS), 0.73 (control vs. WSS), and 0.85 (KS vs. WSS), with corresponding F1 scores of 0.71, 0.72, and 0.83, respectively.
Beyond classification, we apply attention-based visualizations to identify fingerprint regions most salient to model predictions, enhancing interpretability. Together, these findings suggest the presence of syndrome-specific fingerprint features, demonstrating the feasibility of a fingerprint-based artificial intelligence (AI) tool as a noninvasive, interpretable, and accessible future diagnostic aid for the early diagnosis of underdiagnosed genetic syndromes.

\end{abstract}


\section{Introduction}

Kabuki (KS) and Wiedemann-Steiner syndromes (WSS) are two distinct developmental disorders that share overlapping phenotypic features. Clinically, both are characterized by neurodevelopmental delay, growth restriction, and distinctive craniofacial and skeletal features\cite{Adam89, AGGARWAL2017285}. Although genetic testing remains the gold standard for diagnosis, many individuals with KS or WSS remain undiagnosed, particularly in regions where access to genetic testing is limited or unavailable. Abnormal fingerprint characteristics have been observed in both KS\cite{niikawa1982dermatoglyphic} and WSS\cite{sheppard2021expanding}, reflecting disruptions in fetal development that may influence dermatoglyphic formation in utero.
In KS, persistent fetal fingertip pads and atypical ridge patterns have been consistently reported in clinical descriptions. WSS has not been studied as extensively in this context, however clinical observations suggest disruption in normal finger formation, particularly persistence of fetal finger pads\cite{sheppard2022}.

Building on previous work\cite{agustsson2024automated} demonstrating the potential of fingerprint-based image analysis to distinguish individuals with KS from unaffected controls, the present study expands this approach in two key directions: 

$\bullet$ First, we propose a vision transformer classifier and focus on enhancing the interpretability of our approach, providing insights into the specific fingerprint characteristics driving the classification. 

$\bullet$ Second, we examine whether fingerprint-based models can distinguish KS from WSS—two phenotypically overlapping disorders, which are caused by mutations in two highly similar histone methyltransferases (KMT2A and KMT2D). Specifically, since both syndromes are known to be associated with persistent fetal finger pads, we could test whether this anatomical feature is driving the classification or alternatively, whether specific fingerprint features distinguish the two syndromes.

By combining these aims, we seek to demonstrate the potential of a fingerprint-based artificial intelligence (AI) tool as a noninvasive, accessible tool for early and accurate diagnosis of KS and WSS. Access to such a tool would expand the reach of genetic diagnosis to underserved areas lacking sufficient genetic expertise and testing resources, in addition to expanding the patient base for therapeutic research and clinical trials.

\section{Method}

\subsection{Data Collection} 

\begin{figure}[t]
    \centering
    \includegraphics[width=0.32\textwidth]{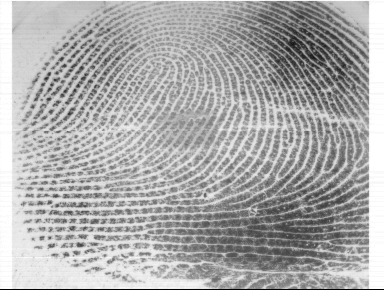}
    \hfill
    \includegraphics[width=0.32\textwidth]{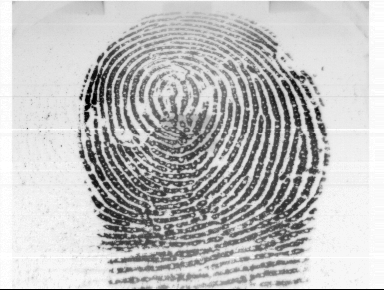}
    \hfill
    \includegraphics[width=0.32\textwidth]{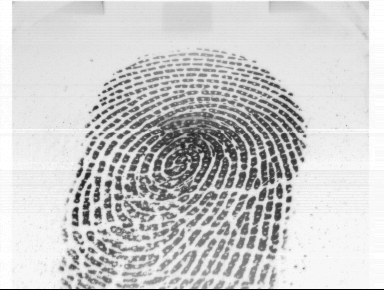}
    \caption{Three representative fingerprint images acquired using the GenePrint mobile application and an external optical scanner. The images correspond to a single finger from patients with KS (left), WSS (center), and a control individual (right)..
}
    \label{fig:three_images}
\end{figure}

\begin{figure}[t]
    \centering
    \includegraphics[width=0.48\textwidth]{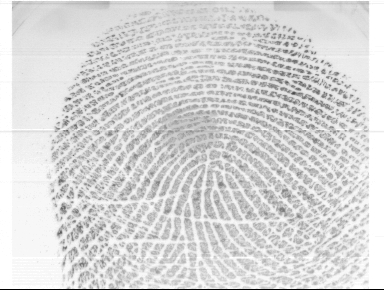}
    \hfill
    \includegraphics[width=0.48\textwidth]{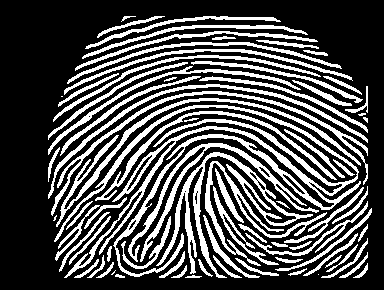}

    \caption{Fingerprint image acquired using the GenePrint mobile application and external optical scanner (left) and the same image after preprocessing with the Gabor filter–based enhancement algorithm (right).
}
    \label{fig:enhancement}
\end{figure}

Participants were recruited across 5 locations between June 2020 and November 2024. The cohort consisted of 75 individuals with a diagnosis of KS, 38 with a diagnosis of WSS, and 120 control individuals. Informed consent was obtained from all participants under protocols approved by the Institutional Review Board of the Johns Hopkins Hospital (NA\_00079185) or the ethics board of Landspitali University Hospital (VRN\_211118) and the National Bioethics Committee (VSN\_21-228).
Data collection was performed using a custom-built Android application called GenePrint, developed to guide and standardize the fingerprint image acquisition process. Fingerprints were captured using an external HID DigitalPersona 4500 optical scanner connected to a smartphone. A visual interface within the app guided the user through the image capture process and a standardized protocol was developed to ensure consistency across data collectors. For each participant, ten fingerprint images, one per finger, were collected. Three representative samples can be seen in Figure~\ref{fig:three_images}.
Raw fingerprint images were converted to 8-bit RGB PNG format, inverted, and processed using an enhancement algorithm that convolves the images with a series of Gabor filters to reduce noise introduced by variable impression artifacts (Figure~\ref{fig:enhancement})\cite{Hong}.
We filtered out low-quality samples using a quality score assigned by the NIST Fingerprint Image Quality 2 (NFIQ 2) software from the National Institute of Standards and Technology (NIST)\cite{nist}. Fingerprints with a NFIQ2 score below 2 were excluded from the analysis, resulting in the removal of 221 low-quality images. After filtering, a total of 2109 fingerprint images were retained for subsequent experiments.

\subsubsection{Model Architecture}

\begin{figure}[t]
    \centering
    \includegraphics[width=1\textwidth]{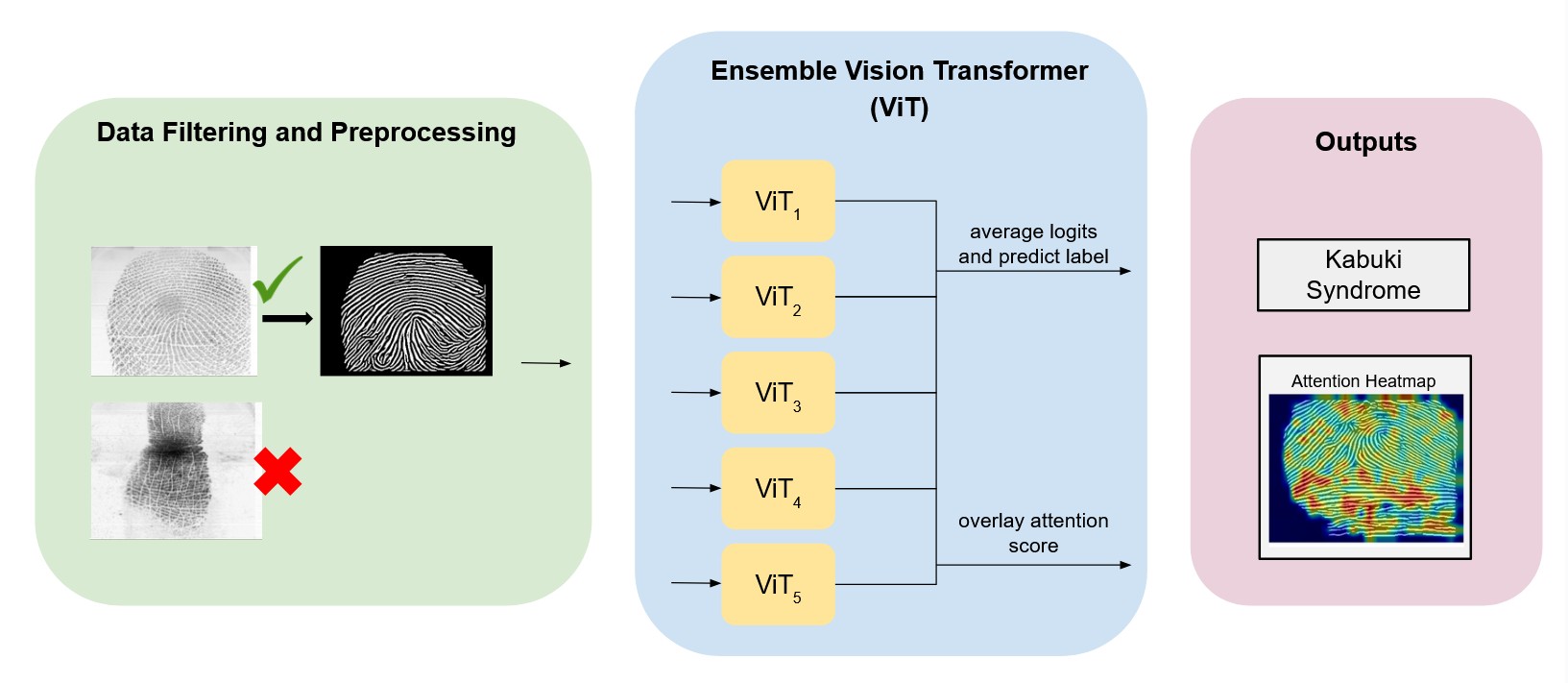}
    \caption{Overview of the fingerprint-based classification pipeline. Fingerprint data were collected from participants with Kabuki syndrome (KS), Wiedemann-Steiner syndrome (WSS), and unaffected controls using a standardized mobile app and an optical fingerprint scanner. Fingerprint images were subsequently preprocessed and filtered according to image quality. A Vision Transformer model was trained and evaluated across three classification tasks using ensemble predictions. Attention heatmaps were generated to visualize model focus and enhance interpretability.}
    \label{pipeline}
\end{figure}
We employed a Vision Transformer (ViT) model for fingerprint image classification. Input images were resized to 224×224 pixels and divided into non-overlapping 16×16 patches, each linearly projected into an embedding of dimension 512 (Control vs. KS; Control vs. WSS) or 256 (KS vs. WSS). A learnable class token was prepended to the sequence to aggregate global image information, and fixed positional encodings were added to retain spatial information across patches. The model architecture comprised three transformer encoder blocks with four self-attention heads and feedforward layers of hidden dimension 1024 (Control vs. KS; Control vs. WSS) or 512 (KS vs. WSS). The output corresponding to the class token was passed to a linear classification head. To improve robustness and generalization, we trained five independent instances of the model for 10 epochs each and ensembled predictions by averaging their output logits. Models were optimized using the Adam optimizer with a learning rate of $3\times10^{-4}$ and no dropout. Larger embedding and hidden dimensions were chosen for disease-control classification models to account for higher sample diversity of the control class, whereas lower dimensional representations were sufficient for KS-WSS discrimination.
To visualize model focus during classification, we generated attention heatmaps by extracting self-attention weights from the class token to image patches. Attention was averaged across all layers and heads to produce a single score per patch. These scores were reshaped to match the patch grid, normalized, and upsampled to the original image size. The resulting attention maps were overlaid on the grayscale fingerprint images to highlight the regions that most influenced the model’s prediction. An overview of the full classification pipeline, including data acquisition, model training, and interpretability via attention heatmaps, is shown in Figure~\ref{pipeline}.

\subsection{Experiments}
We conducted three classification experiments to evaluate the ability of our model to distinguish between (1) control vs. KS, (2) control vs. WSS, and (3) KS vs. WSS. Each participant provided up to 10 fingerprint images, yielding a total of 2330 images. To prevent data leakage, all dataset splits were performed at the participant level, ensuring that fingerprints from the same individual did not appear in both the training and testing sets nor across different cross-validation folds.
For each experiment, we performed an 80/20 split into training and testing sets. To improve robustness and reduce variance due to limited training data, we trained five independent instances of each model for 10 epochs using 5-fold cross-validation, sequentially leaving out each fold as the validation set, and generated final predictions by averaging the output logits across the five trained models. For evaluation, ensemble predictions were used to generate image-level predictions for each fingerprint in the held-out test set, and performance was quantified by aggregating results across all test images. 


\section{Results}

\begin{table}[t]
\centering
\caption{Performance metrics for the three classification tasks.}
\label{tab:performance_metrics}
\begin{tabular}{lccccc}
\toprule
\textbf{Experiment} & \textbf{Accuracy} & \textbf{Precision} &\textbf{Recall} & \textbf{F1 Score} & \textbf{AUC} \\
\midrule
Control vs. KS        & 0.72 & 0.71 & 0.70 & 0.71 & 0.80 \\
Control vs. WSS & 0.80 & 0.73 & 0.73 & 0.72 & 0.73 \\
KS vs. WSS                             & 0.88 & 0.84 & 0.82 & 0.83 & 0.85 \\
\bottomrule
\label{results}
\end{tabular}
\end{table}

\begin{figure}[t]
    \centering
    \includegraphics[width=1\linewidth]{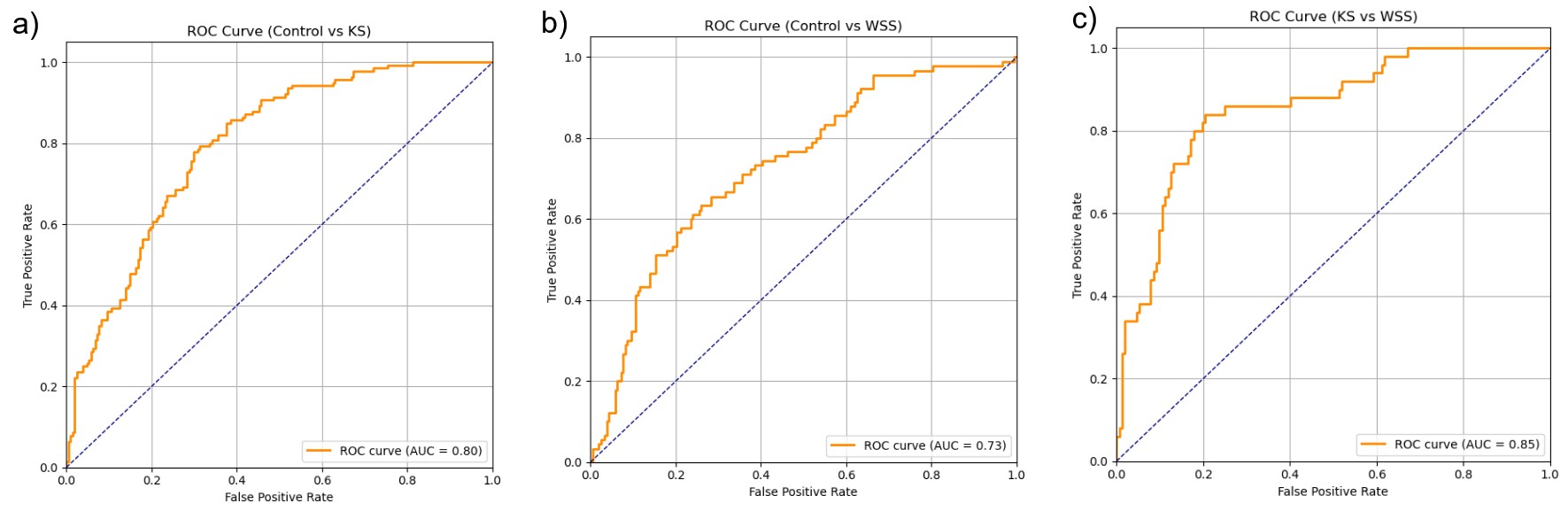}
    \caption{Receiver Operating Characteristic (ROC) curves for the three classification tasks, generated from ensemble predictions on the held-out test set: (a) control vs. Kabuki syndrome (KS) (AUC = 0.80), (b) control vs. Wiedemann-Steiner syndrome (WSS) (AUC = 0.73), and (c) KS vs. WSS (AUC = 0.85).}
    \label{roc}
\end{figure}

\begin{figure}[t]
    \centering
    \includegraphics[width=1\linewidth]{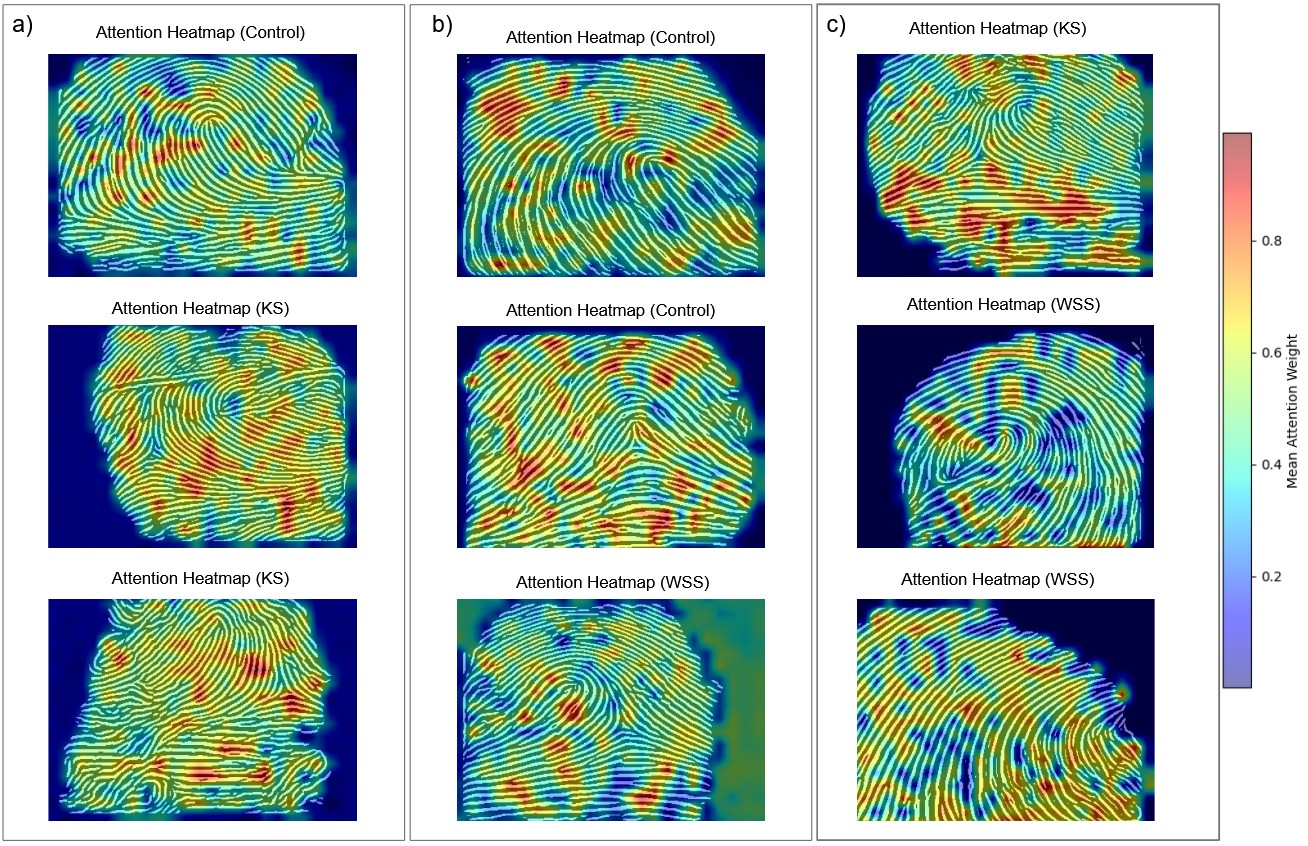}
    \caption{Attention heatmaps generated by the model for the three classification tasks: (a) control vs. Kabuki syndrome (KS), (b) control vs. Wiedemann-Steiner syndrome (WSS), and (c) KS vs. WSS. Each column displays three representative fingerprint images overlaid with attention heatmaps from participants with the corresponding conditions. Column (a) shows fingerprints from one control participant and two participants with KS; column (b) shows fingerprints from two control participants and one participant with WSS; and column (c) shows fingerprints from one participant with KS and two participants with WSS. Heatmaps were derived from class-token self-attention weights, averaged across all transformer layers and heads, and overlaid on the original fingerprint images, with warmer colors indicating higher attention scores.}
    \label{heatmaps}
\end{figure}
We evaluated model performance across three binary classification tasks: (1) distinguishing control individuals from those with KS, (2) distinguishing control individuals from those with WSS, and (3) distinguishing KS from WSS. We report the area under the receiver operating characteristic curve (AUC, see Figure~\ref{roc}), along with accuracy, precision, recall, and F1 score (see Table~\ref{results}). To enhance interpretability, we visualize the model’s attention by generating attention heatmaps by extracting self-attention weights from the class token to individual image patches. Attention weights are averaged across all transformer layers and heads, normalized, and upsampled to the original image resolution, and the resulting heatmaps are overlaid on the corresponding fingerprint images for visualization. The attention heatmaps are shown for representative test samples from each classification task in Figure~\ref{heatmaps}, highlighting the fingerprint regions most influential to the classification.





\section{Conclusion and Discussion}

We introduce a novel vision transformer-based method for identifying KS and WSS from fingerprint images. This pilot study provides initial evidence supporting the potential of AI-driven fingerprint image analysis as a complementary, fast, and noninvasive diagnostic aid for rare genetic diseases. Emphasis on interpretability through attention maps contributes to understanding the unique fingerprint traits associated with each syndrome. Beyond the established presence of persistent fetal fingertip pads, our results suggest the existence of subtle, syndrome-specific fingerprint features that can be captured through deep learning-based image processing and analysis. Notably, although both syndromes are characterized by persistent fetal pads, our attention-based visualizations indicate that diagnostic features extend beyond these regions, suggesting that dermatoglyphic differences are not solely attributable to fetal pad persistence. 
Collectively, these findings provide a foundation for further research into accessible, fingerprint-based computer-aided diagnosis of these underrecognized genetic disorders.
Future research could explore whether fingerprint images can be captured using smartphone cameras instead of an optical scanner, thereby increasing accessibility of the diagnostic tool through smartphones.

\section{Acknowledgements}
We would like to thank participants for participating in this study and family groups for facilitating this study. The collection of fingerprint data was partially funded by the Louma G. Foundation (H.T.B). This research was also supported by Vanderbilt VALIANT Reach, the VALIANT's international exchange program.



\bibliographystyle{spiebib} 
\bibliography{report.bib}

@article{agustsson2024automated,
  author       = {Agustsson, V. I. and Bjornsson, P. A. and Fridriksdottir, A. and Bjornsson, H. T. and Ellingsen, L. M.},
  title        = {Automated fingerprint analysis as a diagnostic tool for the genetic disorder Kabuki syndrome},
  journal      = {Genet Med Open},
  year         = {2024},
  month        = {Aug},
  day          = {7},
  volume       = {2},
  pages        = {101884},
  doi          = {10.1016/j.gimo.2024.101884},
  pmid         = {39669635},
  pmcid        = {PMC11613772}
}

@article{niikawa1982dermatoglyphic,
  author       = {Niikawa, N. and Kuroki, Y. and Kajii, T.},
  title        = {The dermatoglyphic pattern of the Kabuki make-up syndrome},
  journal      = {Clinical Genetics},
  year         = {1982},
  month        = {May},
  volume       = {21},
  number       = {5},
  pages        = {315--320},
  doi          = {10.1111/j.1399-0004.1982.tb01378.x},
  pmid         = {7116676}
}

@article {Adam89,
	author = {Adam, Margaret P and Banka, Siddharth and Bjornsson, Hans T and Bodamer, Olaf and Chudley, Albert E and Harris, Jaqueline and Kawame, Hiroshi and Lanpher, Brendan C and Lindsley, Andrew W and Merla, Giuseppe and Miyake, Noriko and Okamoto, Nobuhiko and Stumpel, Constanze T and Niikawa, Norio},
	title = {Kabuki syndrome: international consensus diagnostic criteria},
	volume = {56},
	number = {2},
	pages = {89--95},
	year = {2019},
	doi = {10.1136/jmedgenet-2018-105625},
	publisher = {BMJ Publishing Group Ltd},
	issn = {0022-2593},
	URL = {https://jmg.bmj.com/content/56/2/89},
	eprint = {https://jmg.bmj.com/content/56/2/89.full.pdf},
	journal = {Journal of Medical Genetics}
}

@article{sheppard2021expanding,
author       = {Sheppard, S. E. and Campbell, I. M. and Harr, M. H. and Gold, N. and Li, D. and et al.},
title        = {Expanding the genotypic and phenotypic spectrum in a diverse cohort of 104 individuals with Wiedemann-Steiner syndrome},
  journal      = {American Journal of Medical Genetics Part A},
  year         = {2021},
  volume       = {185},
  number       = {6},
  pages        = {1649--1665},
  month        = {Jun},
  doi          = {10.1002/ajmg.a.62124},
  pmid         = {33783954},
  pmcid        = {PMC8631250},
  note         = {Erratum in: Am J Med Genet A. 2022 Mar;188(3):1015. doi: 10.1002/ajmg.a.62567}
}

@article{AGGARWAL2017285,
title = {Wiedemann-Steiner syndrome: Novel pathogenic variant and review of literature},
journal = {European Journal of Medical Genetics},
volume = {60},
number = {6},
pages = {285-288},
year = {2017},
issn = {1769-7212},
doi = {https://doi.org/10.1016/j.ejmg.2017.03.006},
url = {https://www.sciencedirect.com/science/article/pii/S1769721216302440},
author = {Anjali Aggarwal and David F. Rodriguez-Buritica and Hope Northrup}
}

@ARTICLE{hong,
  author={Lin Hong and Yifei Wan and Jain, A.},
  journal={IEEE Transactions on Pattern Analysis and Machine Intelligence}, 
  title={Fingerprint image enhancement: algorithm and performance evaluation}, 
  year={1998},
  volume={20},
  number={8},
  pages={777-789},
  doi={10.1109/34.709565}}

@misc{nist,
  author = {Elham Tabassi and Martin Olsen and Oliver Bausinger and Christoph Busch and Andrew Figlarz and Gregory Fiumara and Olaf Henniger and Johannes Merkle and Timo Ruhland and Christopher Schiel and Michael Schwaiger},
  title = {NIST Fingerprint Image Quality 2},
  year = {2021},
  month = {2021-07-13 04:07:00},
  publisher = {NIST Interagency/Internal Report (NISTIR), National Institute of Standards and Technology, Gaithersburg, MD},
  url = {https://tsapps.nist.gov/publication/get_pdf.cfm?pub_id=920087},
  doi = {https://doi.org/10.6028/NIST.IR.8382},
  language = {en},
}

@incollection{sheppard2022,
  author       = {Sheppard, Sarah E. and Quintero-Rivera, Fabiola},
  title        = {Wiedemann-Steiner Syndrome},
  booktitle    = {GeneReviews{\textregistered}},
  editor       = {Adam, Margaret P. and Feldman, Jack and Mirzaa, Ghayda M. and others},
  year         = {2022},
  month        = {May 26},
  publisher    = {University of Washington, Seattle},
  address      = {Seattle (WA)},
  note         = {Available from: \url{https://www.ncbi.nlm.nih.gov/books/NBK580718/}},
  series       = {GeneReviews},
  url          = {https://www.ncbi.nlm.nih.gov/books/NBK580718/}
}

\end{document}